\def\paperlanguage{}
\pdfoutput=1 

\documentclass[letterpaper, 10 pt, conference]{ieeeconf}  

\usepackage{bm}
\usepackage{cite}
\include{preamble}

\IEEEoverridecommandlockouts                              

\title{\LARGE \textbf
  {
    \switchlanguage%
    {%
      A Method of Joint Angle Estimation Using Only Relative Changes in Muscle Lengths for Tendon-driven Humanoids\\with Complex Musculoskeletal Structures
    }%
    {%
      筋骨格ヒューマノイドの肩甲骨や多関節筋を含む複雑な骨格における相対筋長のみを用いた関節角度推定手法
    }%
  }
}

\author{Kento Kawaharazuka, Shogo Makino, Masaya Kawamura, Yuki Asano, Kei Okada and Masayuki Inaba
  \thanks{The authors are with the Department of Mechano-Informatics, Graduate School of Information Science and Technology, The University of Tokyo, 7-3-1 Hongo, Bunkyo-ku, Tokyo, 113-8656, Japan.
    {\texttt\small [kawaharazuka, makino, kawamura, asano, k-okada, inaba]@jsk.t.u-tokyo.ac.jp}
  }
}
\begin{document}

\maketitle
\thispagestyle{empty}
\pagestyle{empty}

\begin{abstract}
  \switchlanguage%
  {%
    Tendon-driven musculoskeletal humanoids typically have complex structures similar to those of human beings, such as ball joints and the scapula, in which encoders cannot be installed.
    Therefore, joint angles cannot be directly obtained and need to be estimated using the changes in muscle lengths.
    In previous studies, methods using table-search and extended kalman filter have been developed.
    These methods express the joint-muscle mapping, which is the nonlinear relationship between joint angles and muscle lengths, by using a data table, polynomials, or a neural network.
    However, due to computational complexity, these methods cannot consider the effects of polyarticular muscles.
    In this study, considering the limitation of the computational cost, we reduce unnecessary degrees of freedom, divide joints and muscles into several groups, and formulate a joint angle estimation method that takes into account polyarticular muscles.
    Also, we extend the estimation method to propose a joint angle estimation method using only the relative changes in muscle lengths.
    By this extension, which does not use absolute muscle lengths, we do not need to execute a difficult calibration of muscle lengths for tendon-driven musculoskeletal humanoids.
    Finally, we conduct experiments in simulation and actual environments, and verify the effectiveness of this study.
  }%
  {%
    筋骨格ヒューマノイドは一般的に球関節や肩甲骨の存在からエンコーダ等を入れることができず、関節角度値を直接得ることが出来ないため筋長変化から関節角度を推定する。
    そして、現在までにテーブルサーチや拡張カルマンフィルタを用いた手法が開発されてきた。
    そしてこの際には、関節空間と筋空間の非線形な写像である関節-筋空間マッピングを、データテーブルや多項式、ニューラルネットワークによって表現してきた。
    しかしこれらは計算量的観点から多関節筋の影響を考慮しておらず、正しいとは言えない。
    我々はこの際の計算量的な限界から、肩甲骨の自由度の削減、上肢の関節群と筋群のグループ分け、多関節を含む場合の関節角度推定の定式化を試みる。
    また、それをより発展させ、相対的な筋長変化のみから関節角度推定を行う方法を開発する。
    これによって、絶対的な筋長を用いないため、筋長のキャリブレーションを行う必要がなくなると考える。
    そのためのpreliminaryな実験をシミュレーションと実機で行い、その有用性と、これからの関節角度推定について議論を行う。
  }%
\end{abstract}

\section{INTRODUCTION} \label{sec:1}
\begin{figure}[t]
  \centering
  \includegraphics[width=0.75\columnwidth]{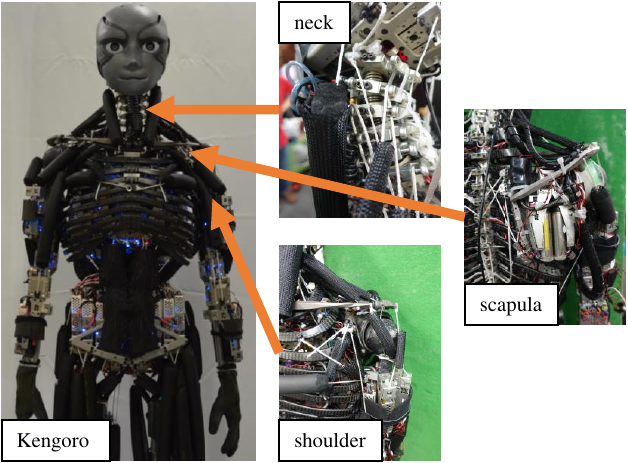}
  \caption{Complexity of tendon-driven musculoskeletal humanoids. This figure shows that the neck and shoulder girdle of Kengoro have very complex structures like human beings.}
  \label{figure:kengoro-overview}
  \vspace{-1.0ex}
\end{figure}

\switchlanguage%
{%
  Tendon-driven musculoskeletal humanoids \cite{ijars2013:nakanishi:approach, artl2013:wittmeier:ecce, humanoids2013:michael:anthrob, humanoids2016:asano:kengoro}, which imitate not only the proportion but also the bone and muscle structures of human beings, have been developed vigorously.
  They are expected to play an active part in the future because of their contact softness, variable stiffness, flexible spine, etc.
  We can apply the human reflex or sensor system to these humanoids and make use of the obtained results for human beings.

  As shown in \figref{figure:kengoro-overview}, tendon-driven musculoskeletal humanoids have ball joints like the shoulder and scapula, and structures with multiple degrees of freedom (multi-DOFs) like the spine and neck, etc., in which encoders cannot be installed.
  Therefore, their joint angles cannot be directly measured by encoders like in ordinary axis-driven humanoids, and joint angles need to be estimated from the changes in muscle lengths.
  On the other hand, there are studies to directly estimate angles of ball joints using cameras \cite{icra2006:urata:joint}, but we do not consider them in this study.
  In previous studies, methods using table-search \cite{icra2010:nakanishi:table}, extended kalman filter (EKF) with polynomial expression \cite{humanoids2015:okubo:muscle-learning}, and the EKF with neural network (NN) \cite{ral2018:kawaharazuka:vision-learning} have been proposed.
  In these methods, we invariably need joint-muscle mapping (JMM), which expresses the nonlinear relationship between joint angles and muscle lengths.
  Usually, we generate a data set of joint angles and their corresponding muscle lengths, and train JMM with it.
  However, there is a problem in terms of computational complexity.
  Tendon-driven musculoskeletal humanoids have numerous polyarticular muscles because they imitate human beings.
  Therefore, for example, when we express muscle lengths using polynomials of joint angles, the larger the number of polyarticular muscles that cross joints are, the harder the polynomial regression becomes computationally.
  There are similar problems in the methods using a data table and neural network.
  In particular, to construct JMM of the whole body is realistically difficult.
  To solve this problem, we reduced unnecessary DOFs, divided joints and muscles into several groups, and formulated a joint angle estimation method using the JMM that includes polyarticular muscles.
  By these methods, we solved the problem of computational cost and realized stable joint angle estimation.
}%
{%
  近年、人間の体のプロポーションだけでなく、人間の筋構造や骨格構造までも模倣した筋骨格ヒューマノイドの開発が盛んである\cite{ijars2013:nakanishi:approach, artl2013:wittmeier:ecce, humanoids2013:michael:anthrob, humanoids2016:asano:kengoro}。
  筋骨格ヒューマノイドは可変剛性や接触のやわらかさ、構造的な自由度の多さから今後の活躍が期待されている。
  また、人間を理解するという意味でも非常に有用であり、筋骨格ヒューマノイドで得た知見を人間に適用すること、そして、新たなロボットの開発に活かすことができると考える。

  筋骨格ヒューマノイドは\figref{figure:kengoro-overview}のような複雑な肩甲骨や肩の球関節、多くの椎等で構成される首などがあるため、通常のヒューマノイドのようにエンコーダを関節に入れることができない.
  そのため, 直接関節角度値を得ることは出来ず、筋長の変化から関節角度を推定する。
  一方で, モーションキャプチャやカメラと光ファイバーを用いた関節角度推定手法も存在するが, 本研究ではこれらについては扱わない.
  関節角度の推定方法として、現在までにテーブル探索による方法\cite{icra2010:nakanishi:table}や拡張カルマンフィルタと多項式近似による方法\cite{humanoids2015:okubo:muscle-learning}、拡張カルマンフィルタとニューラルネットワークによる方法\cite{ral2018:kawaharazuka:vision-learning}が提案されている。
  これらの方法には必ず、関節空間から筋空間への非線形性な写像である関節-筋長マッピングが必要となる。
  関節角度と筋長の対応のデータセットを作成し、それを用いて関節-筋空間マッピングを得るが、ここには計算量的観点から大きな問題が生じる。
  筋骨格ヒューマノイドは人間を模しているが故に、多くの多関節筋が存在する。
  それゆえに、例えばある筋の長さを関節角の多項式近似で表す場合は、多くの関節を跨ぐ筋であればあるほど巨大な多項式による近似となってしまい、計算量が膨大になる。
  これはテーブル探索やニューラルネットワークによる方法でも同様であり、データ作成の計算量は莫大な数になる。
  特に、全身の筋と全身の関節角の非線形な関係を表そうとすることは現在では無謀であると思われる。
  このような問題に対して我々は、自由度の削減、関節群と筋群のグルーピング、多関節筋を含む場合の関節角度推定の定式化を行い、計算量的な問題をクリアし、安定した関節角度推定を実現させる。
}%

\begin{figure*}[t]
  \centering
  \includegraphics[width=1.9\columnwidth]{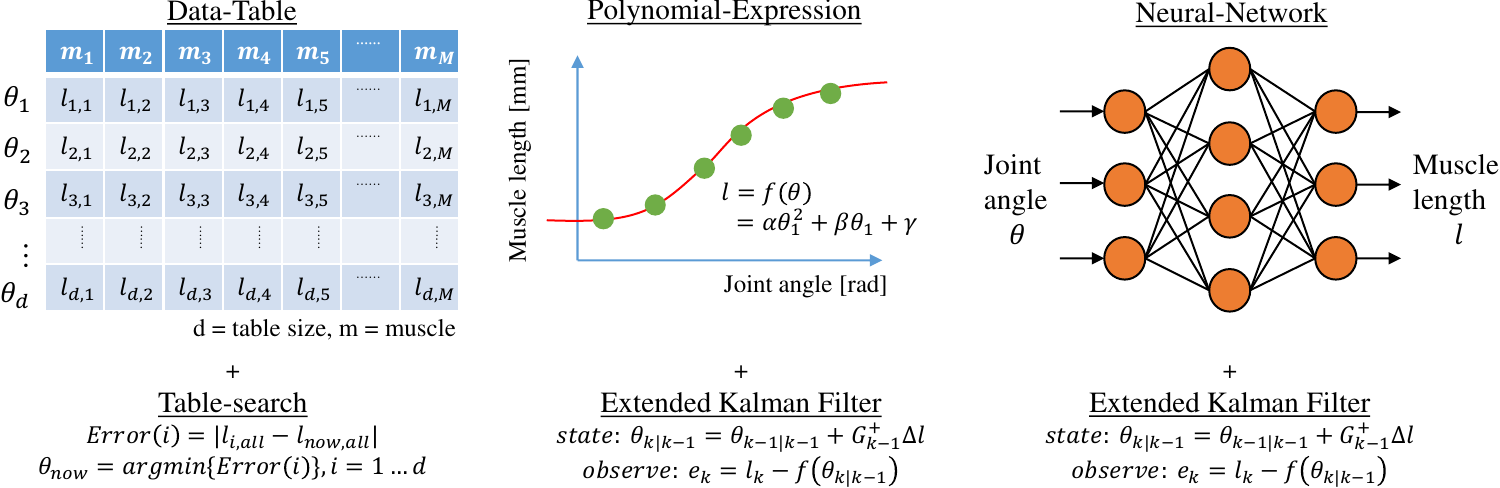}
  \caption{Overview of the joint angle estimation methods: table-search \cite{icra2010:nakanishi:table}, EKF with polynomial regression \cite{humanoids2015:okubo:muscle-learning}, and EKF with neural network \cite{ral2018:kawaharazuka:vision-learning}.}
  \label{figure:various-joint-angle-estimation}
  \vspace{-1.0ex}
\end{figure*}

\switchlanguage%
{%
  Also in this study, we propose a new joint angle estimation method using not absolute muscle lengths but only the relative changes in muscle lengths.
  In a previous study, Nakanishi, et al. \cite{icra2010:nakanishi:table} developed a joint angle estimation method using only the relative changes in muscle lengths through table-search.
  However, this method required a large computational cost and was difficult to apply when the number of DOFs was large.
  Thus, in this study, we extend the method of Ookubo, et al. \cite{humanoids2015:okubo:muscle-learning} and show that the joint angle estimation using only the relative changes in muscle lengths is possible.
  This method makes the calibration of muscle lengths, which is difficult for tendon-driven musculoskeletal humanoids, unnecessary, and the self-body posture can be obtained during exercise.
  This study can be applied to robots using tendon structures such as the ACT Hand \cite{icra2004:vande:acthand}, the tensegrity robot \cite{tro2006:paul:design}, and exoskeletons \cite{yan2015review}.
  Also, we hope that this is one step for tendon-driven musculoskeletal humanoids to catch up to ordinary axis-driven humanoids.

  This paper is organized as follows.
  In \secref{sec:1}, we stated the motivation and goal of this study.
  In \secref{sec:2}, we will summarize the previous joint angle estimation methods and clarify their problems.
  In \secref{sec:3}, to solve these problems, we will discuss the reduction of unnecessary DOFs, division of joints and muscles into several groups, and formulation of a joint angle estimation method that includes polyarticular muscles.
  In \secref{sec:4}, we will extend the previous joint angle estimation method and propose a new joint angle estimation method using only the relative changes in muscle lengths.
  In \secref{sec:5}, we will conduct experiments using these proposed methods in simulation and actual environments, and verify their effectiveness.
  Finally, in \secref{sec:6}, we will state the conclusion and future works.
}%
{%
  またこれらを発展させ、本研究では、絶対的な筋長からでなく相対的な筋長の変化のみから関節角度推定を行う手法を開発する。
  中西らはテーブルサーチを用いた方法によって、相対的な筋長のみから関節角度を推定する手法を開発した\cite{icra2010:nakanishi:table}。
  しかしこの方法では自由度が多い場合には適用が難しい。
  そこで本研究では、大久保らの手法\cite{humanoids2015:okubo:muscle-learning}の自然な拡張によって相対的な筋長変化のみから関節角度推定が行えることを示す。
  これによって、非常に難しい筋長のゼロ点キャリブレーションする必要がなくなり、体を動かしているうちに自然と自身の身体状態が得られるようになると考える。
  本研究は筋骨格ヒューマノイドだけではなく, 腱で動くようなACT HAND \cite{icra2004:vande:acthand}や, Tensegrity Robot \cite{tro2006:paul:design}にも関係すると考える.
  そして、筋骨格ヒューマノイドが軸駆動型ヒューマノイドに追いつく一歩となることを願う。

  本研究は以下のような構成となっている。
  \secref{sec:1}では本論文の動機と目標について述べた。
  \secref{sec:2}では現在までに行われてきた筋長からの関節角度推定をまとめ、問題点等を洗い出す。
  \secref{sec:3}では肩甲骨自由度の削減、関節群と筋群のグルーピング、多関節筋を含む関節-筋空間マッピングに対応した関節角度推定の定式化について説明する。
  \secref{sec:4}では相対的な筋長のみを用いた関節角度推定の、大久保らの関節角度推定からの自然な拡張を行う。
  \secref{sec:5}では多関節を含むような関節角度推定・相対的な筋長のみからの関節角度推定についてシミュレーションと実機でそれぞれ実験を行い本手法の有効性を示し、議論を行う。
  最後に、\secref{sec:6}では結論と今後の方針について述べる。
}%

\section{Joint Angle Estimation Methods in Tendon-driven Musculoskeletal Humanoids and Their Problems} \label{sec:2}
\switchlanguage%
{%
  First, we will summarize previous studies on joint angle estimation for tendon-driven musculoskeletal humanoids.
  Next, we will state the problems on the construction of JMM, which is invariably necessary for these estimation methods.
  Also, we will consider a joint angle estimation method that does not use the absolute muscle lengths but only the relative changes in muscle lengths.
}%
{%
  初めに、現在までに開発されてきた筋骨格ヒューマノイドにおける関節角度推定をまとめる。
  次に、その際に必ず必要となる関節-筋空間マッピングに関する問題点を挙げる。
  また、絶対的な筋長からの関節角度推定だけでなく、相対的な筋長のみを使った関節角度推定に関する考察も行う。
}%

\subsection{Joint Angle Estimation in Tendon-driven Musculoskeletal Humanoids}
\switchlanguage%
{%
  Because encoders cannot be installed in complex musculoskeletal structures such as ball joints, the scapula, and the spine, we need to estimate joint angles from the changes in muscle lengths.
  In this study, we do not consider the use of a motion capture system, because it is not available for outdoor use and requires large facilities.
  We show the summary of previous joint angle estimation methods in \figref{figure:various-joint-angle-estimation}.

  Nakanishi, et al. developed the method using table-search \cite{icra2010:nakanishi:table}.
  In this method, JMM is constructed as a data table of the joint angles and muscle lengths of a geometric model, and the current joint angles are obtained from the current muscle lengths by using pattern matching.
  In this study, a geometric model is expressed with each muscle connecting the start point, relay points, and end point linearly.
  However, because this method does not use the time series aspect of the data, the continuity of joint angles is not ensured, the precision and computational cost depend on the size of the data table, and it is susceptible to noise.

  Next, Ookubo, et al. proposed the joint angle estimation method using extended kalman filter (EKF) with polynomial regression \cite{humanoids2015:okubo:muscle-learning}.
  In this method, JMM is constructed using polynomials of joint angles to express muscle lengths in the geometric model, and the muscle jacobian is obtained by differentiating the JMM.
  The EKF predicts current joint angles by using the inverse matrix of the muscle jacobian and the relative changes in muscle lengths, and corrects the estimated joint angles by comparing muscle lengths of the actual robot with the estimated muscle lengths, which are calculated from the JMM and predicted joint angles.
  Also, Kawaharazuka, et al. estimated joint angles by replacing the polynomials with a neural network \cite{ral2018:kawaharazuka:vision-learning}.
  The benefit of using a neural network is that the data structure is suitable for the update of JMM.
  The state and observation equation of the EKF are shown below,
  \begin{align}
    \bm{\theta}_{k|k-1} &= \bm{\theta}_{k-1|k-1} + G^{+}(\bm{\theta}_{k-1|k-1})\delta\bm{z}_k \label{equation:state-normal} \\
    \bm{z}_{k} &= f(\bm{\theta}_{k|k-1}) \label{equation:observation-normal}
  \end{align}
  where $\bm{\theta}$ is joint angles, $G$ is muscle jacobian, $G^{+}$ is a pseudo inverse matrix of $G$, $\delta\bm{z}$ is the relative changes in muscle lengths of the actual robot, $\bm{z}$ is absolute muscle lengths of the actual robot, and $f$ is JMM.

  It is essential that all estimation methods use JMM, stated above, and the construction of JMM greatly affects their computational complexity and precision.
  \begin{align}
    \bm{l} = f(\bm{\theta})
  \end{align}

}%
{%
  筋骨格ヒューマノイドは肩の球関節や肩甲骨・背骨等の複雑な構造を持つがゆえに関節角度値を直接エンコーダ等から求めることができないため、筋長の変化から関節角度を推定する。
  注意として、本研究では屋外での使用や機材の観点からモーションキャプチャ等の使用は考えない。
  先行研究における関節角度推定をまとめた図を\figref{figure:various-joint-angle-estimation}に示す。

  まず先行研究として、中西らはテーブルサーチを用いた方法を開発した\cite{icra2010:nakanishi:table}。
  これは、幾何モデルを用いて関節-筋空間マッピングを単純なテーブルとして保持しておき、実機での、ある筋長のときの関節角度をパターンマッチングを用いて得るというものである。
  しかし、この方法は時系列関係を保持しないため関節角度の連続性は保証されず、精度もデータテーブルの大きさに依存し、ノイズにも弱い。
  そのため、基本的に本研究ではこの後説明する手法をベースとする。

  次に、大久保らの、多項式近似と拡張カルマンフィルタを用いた関節角度推定が挙げられる\cite{humanoids2015:okubo:muscle-learning}。
  これは、幾何モデルを用いて、筋長を関節角度の多項式として得ておき、それを微分することで筋長ヤコビアンを得る。
  そして筋長ヤコビアンの逆行列と実機の筋長変化を用いることで関節角度を予測し、その予測した関節角度を関節-筋空間マッピングによって筋長に変換した際の値と実機の絶対筋長を比較することで関節角度の補正を行うカルマンフィルタを実行する。
  また、河原塚らはこの際の多項式近似をニューラルネットワークに置き換えることで関節角度推定を行っている\cite{ral2018:kawaharazuka:vision-learning}。
  ニューラルネットワークを使うことの利点は、オンラインでの関節-筋空間マッピングの更新に向いていることである。
  拡張カルマンフィルタの式は以下のようになっている。
  \begin{align}
    \nonumber\mbox{Predict}\;\;\;\;\;\;\;\;\;\;\;\;& \\
    \bm{\theta}_{k|k-1} &= \bm{\theta}_{k-1|k-1} + G^{+}(\bm{\theta}_{k-1|k-1})\delta\bm{z}_k \label{equation:state-normal} \\
    P_{k|k-1} &= P_{k-1|k-1} + Q \\
    \nonumber\mbox{Update}\;\;\;\;\;\;\;\;\;\;\;\;& \\
    \bm{e}_{k} &= \bm{z}_{k} - f(\bm{\theta}_{k|k-1}) \label{equation:observation-normal} \\
    G_{k} &= \left.\frac{\partial f}{\partial \bm{\theta}}\right|_{\bm{\theta} = \bm{\theta}_{k|k-1}} = G(\bm{\theta}_{k|k-1}) \\
    S_{k} &= G_{k}P_{k|k-1}G^T_{k} + R \\
    K_{k} &= P_{k|k-1}G^T_{k}S^{-1}_{k} \\
    \bm{\theta}_{k|k} &= \bm{\theta}_{k|k-1} + K_{k}e_{k} \\
    P_{k|k} &= (I-K_{k}G_{k})P_{k|k-1} \label{equation:last-normal}
  \end{align}
  ここで、$\bm{\theta}$が関節角度、$G$が筋長ヤコビアン、$G^{+}$が筋長ヤコビアンの擬似逆行列、$\delta\bm{z}$が実機の筋長変化分、$P$が関節角度推定値の誤差の共分散行列、$Q$が時間遷移に関する誤差の共分散行列、$\bm{e}$が観測残差、$\bm{z}$が実機の絶対筋長、$f$が関節-筋空間マッピングを表す関数、$R$が観測空間の共分散行列、$S$が観測残差の共分散行列、$K$がカルマンゲインである。

  ここで重要なのは、どの方法でも関節角度空間から筋長空間への写像である関節-筋空間マッピング
  \begin{align}
    \bm{l} = f(\bm{\theta})
  \end{align}
  が必要であり、これが計算量や精度に非常に効いてくるということである。
}%

\subsection{Problems in Construction of the Joint-Muscle Mapping}
\switchlanguage%
{%
  We showed three joint angle estimation methods using JMM expressed by the data table, polynomials, and NN.
  In this subsection, we will explain the advantages and disadvantages of these structures.

  First, expressions using data tables are superior in terms of simplicity.
  However, when we use data tables, problems occur in terms of computational cost.
  When the number of DOFs of the robot is $D$, and the required precision is of which when the movable ranges of the DOFs are divided into $N$ parts, the size of the data table is $N^D$.
  For example, when the number of DOFs of the shoulder is 3, the range of joint angles is 180 deg, and we need a resolution of 0.1 deg, we need a data table the size of $1800^3$.
  Of course, the more the resolution and the number of DOFs increase, the more the size of the data table increases.
  Therefore, this method is not practical for tendon-driven musculoskeletal humanoids with multi-DOFs like Kengoro \cite{humanoids2016:asano:kengoro}.

  Next, regarding the method using EKF with polynomial regression, because JMM is calculated as polynomials, with a degree of $P$, a smooth mapping can be obtained, JMM can be differentiated analytically, and the precision is better than in the method using a data table.
  Also, this method does not need a large memory like with the data table, and we only need the coefficients of the polynomials.
  However, at the same time, the calculation of the data set construction and polynomial regression require a large computational cost.
  In the data set construction of this method, we divide the movable ranges of joint angles into $N$ parts, move all DOFs, make all combinations, obtain muscle lengths at each posture, and construct a data set the size of $n = N^D$.
  Because polynomial regression can compensate for the values among the data set compared to the method using a data table, the $N$ of this method needs to be only about 6--9.
  When the number of muscles and DOFs included in the JMM is $M$ and $D$, respectively, we solve the polynomial regression to calculate $C$ as shown below and the computational complexity to calculate it is $O({(N^D)^2} _{D+1}H_{P})$ ($H$ means combination with repetition).
  \begin{align}
  \begin{pmatrix} l_{1}^{1}&\cdots&l_{M}^{1} \\ \vdots&\ddots&\vdots \\ l_{1}^{n}&\cdots&l_{M}^{n} \end{pmatrix} = C\begin{pmatrix} \prod_{i=\{0\dots0\}}^{N}\theta_{i}^{1}&\cdots&\prod_{i=\{D \dots D\}}^{N}\theta_{i}^{1} \\ \vdots&\ddots&\vdots \\ \prod_{i=\{0\dots0\}}^{N}\theta_{i}^{n}&\cdots&\prod_{i=\{D \dots D\}}^{N}\theta_{i}^{n} \end{pmatrix}
  \end{align}
  In this equation, $l_{i}^{j}$ is the $i$-th muscle of the $j$-th data, $\theta_{i}^{j}$ is the $i$-th DOF of the $j$-th data, and $\theta_0$ is $1$ for simplicity.
  For example, when $N$ is 8 and $D$ is 10, the size of the data set is about $1E+9$ (to generate each data takes about 10 msec) and to construct the data set takes a long time.
  When $P$ is 5, the computational cost to calculate the polynomial regression is about $1E+21$, and it is realistically difficult.

  Finally, the computational cost of the method using a NN is almost the same as in the method using polynomials.
  A bottleneck of the computational cost occurs when constructing the data set and training a NN with it.

  From these considerations, we need to group joints and muscles into small parts, and construct JMM regarding each body part.
  At the same time, because polyarticular muscles cross JMMs, each JMM includes DOFs duplicated in several JMMs (we call a JMM like this polyarticular-JMM), and we need to cope with this problem.
}%
{%
  関節-筋空間マッピングの表現方法としてデータテーブル・多項式・ニューラルネットワークの3つを挙げたが、これら一つ一つの長所と短所について述べていく。

  まずデータテーブルについてだが、データテーブルを用いる方法は非常に単純であるという点で優れている。
  しかし、データテーブルを用いる場合は計算量に関して大きな問題が生じる。自由度数が$D$、関節角度範囲を$N$個に分割した程度の精度が欲しいとき、テーブルの大きさは$N^D$となる。
  例えば肩の3自由度が角度限界範囲が180度であり、それを1度ずつの分解能で測るとすると$180^3$のテーブルが必要となる。
  当然これは自由度が増えるごとに指数関数的にテーブルが増えるため、多自由度な筋骨格ヒューマノイドに適用するには現実的でないことがわかる。

  次に、多項式近似を用いる方法についてだが、次数$P$の多項式として関節-筋空間マッピングを求めるため、滑らかな写像が求められ、解析的に微分が可能である。
  そのため、精度は非常に良く出る。
  また、データテーブルのように大きなメモリを消費せず、多項式の係数だけ保持すれば良いこととなる。
  しかし同時に、データセット作成と多項式近似を求める際に大きな計算量が必要である。
  データ作成では、関節角度範囲を$N$個に分割し、それぞれの関節を全て動作させた際の筋長を得るため、$N^D$個のデータが作成される。
  ただし、データテーブルとは違い、多項式近似で補間するため、$N$は6--9程度で事足りる。
  そして、多項式の係数を求める際の計算量は、その関節-筋空間マッピングに$M$本の筋肉、$D$自由度が含まれているとすると、$O({(N^D)^2} _{D+1}H_{P})$の計算量が必要である。
  例えば$N$を8、$D$を10とした場合、データセットは約1e9であり、データセットを作成するのに非常に時間がかかる。
  また、同様に$P$を5とした場合、多項式近似を求めるには約1e21の計算量が必要であり、無謀である。

  最後にニューラルネットワークを用いる方法についてだが、この場合も同様にデータセットの作成、学習の時間がボトルネックとなる。

  これらの考察から、関節群筋群をある程度小さなグループに小分けし、それぞれを個別に関節-筋空間マッピングを構成する必要がある。
  そして同時に、その個々の関節-筋空間マッピングを多関節筋が跨ぐため、それぞれの関節-筋空間マッピングに重複した関節が含まれることとなり、それらを適切に処理する必要が生じる。
  詳しくは次章で説明を行う。
}%

\subsection{Joint Angle Estimation Using Only Relative Changes in Muscle Lengths}
\switchlanguage%
{%
  Nakanishi, et al. stated that not only the joint angle estimation method using the absolute muscle lengths (absolute-JAE) but also the estimation method using only the relative changes in muscle lengths (relative-JAE) \cite{icra2010:nakanishi:table} is possible by expressing JMM as a data table.
  This method uses the nonlinear feature of JMM and calculates the current joint angles by ensuring consistency in time series data of muscle lengths.
  However, because the method must scan all the spaces of the data table, the larger the number of DOFs is, the larger the computational cost becomes, so this method cannot be used.
  Therefore, in this study, we extend the joint angle estimation method proposed by Ookubo, et al. \cite{humanoids2015:okubo:muscle-learning}, and try to estimate joint angles using only the relative changes in muscle lengths.
  This method can solve problems in computational cost, precision, and the size of data.

  In this study, the absolute muscle lengths mean the values of muscle lengths when they are calibrated to 0 at the initial posture (all joint angles are 0 deg).
  Although these muscle lengths are relative to those of the initial posture in a sense, we call these muscle lengths absolute muscle lengths for simplicity.
  Also, we assume that each muscle length is measured by an incremental encoder equipped in each muscle motor.
}%
{%
  中西らはテーブルサーチを用い、絶対筋長からの関節角度推定だけでなく、相対的な筋長のみを用いた関節角度推定についても述べている\cite{icra2010:nakanishi:table}。
  これは関節-筋空間マッピングの非線形性を利用しており、筋長の時系列変化に整合性が取れるような初期の関節角度を求める、というものである。
  しかし、多自由度になるほどこの手法ではデータテーブル空間を全て走査しなければならないため、計算量が膨大になってしまい無謀である。
  そこで、本研究では大久保らの関節角度推定\cite{humanoids2015:okubo:muscle-learning}のシンプルな拡張を行うことで、相対的な筋長のみを用いた関節角度推定を行うことを試みる。
}%

\section{Joint Angle Estimation of Complex Musculoskeletal Structures} \label{sec:3}
\switchlanguage%
{%
  In this section, we will explain how we should group joints and muscles and construct each polyarticular-JMM, and how we should formulate the joint angle estimation method using these polyarticular-JMMs, when the robot has many complex structures like human beings, such as the scapula and polyarticular muscles.
  The procedure is shown as below and we will explain each step.
  \begin{enumerate}
    \item reduce unnecessary DOFs of joints to make the size of the JMM small.
    \item divide joints and muscles into several groups, and construct polyarticular-JMMs regarding each group.
    \item change the formulation of the previous joint angle estimation so as to cope with the DOFs that are duplicated in several polyarticular-JMMs.
  \end{enumerate}
}%
{%
  本章では、肩甲骨などの複雑な関節や多関節筋を含む場合に、どのように関節-筋空間マッピングを小分けし、関節角度推定を行うかについて述べる。
  手順としては、
  \begin{enumerate}
    \item 関節-筋空間マッピングの大きさを小さくするため、不必要な関節の自由度を削除する。
    \item 関節群-筋群を小分けし、それぞれのグループに関節-筋空間マッピングを構築する。
    \item 関節角度推定の定式化を変更し、小分けしたグループ間に重複して存在する自由度に処理を施す。
  \end{enumerate}
  のようになっており、一つずつ説明する。
}%

\subsection{Reducing DOFs of the Upper Limb} \label{subsec:reduce-scapula}
\begin{figure}[t]
  \centering
  \includegraphics[width=1.0\columnwidth]{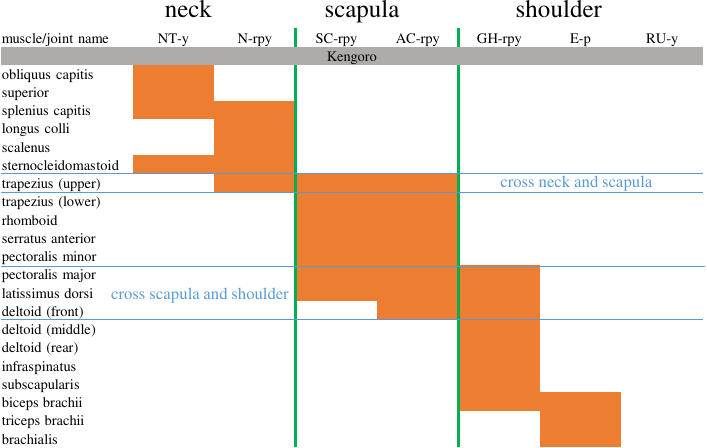}
  \caption{Correspondence between joints and muscles of Kupper limb of Kengoro. The joints that each muscle can move are painted orange. NT is the neck top 1-DOF joint, N is the neck 3-DOF joint, SC is the sternoclavicular joint, AC is the acromioclavicular joint, GH is the glenohumeral joint, E is the elbow joint, and RU is the radioulnar joint.}
  \label{figure:kengoro-joint-muscle-table}
  \vspace{-1.0ex}
\end{figure}

\begin{figure}[htb]
  \centering
  \includegraphics[width=0.9\columnwidth]{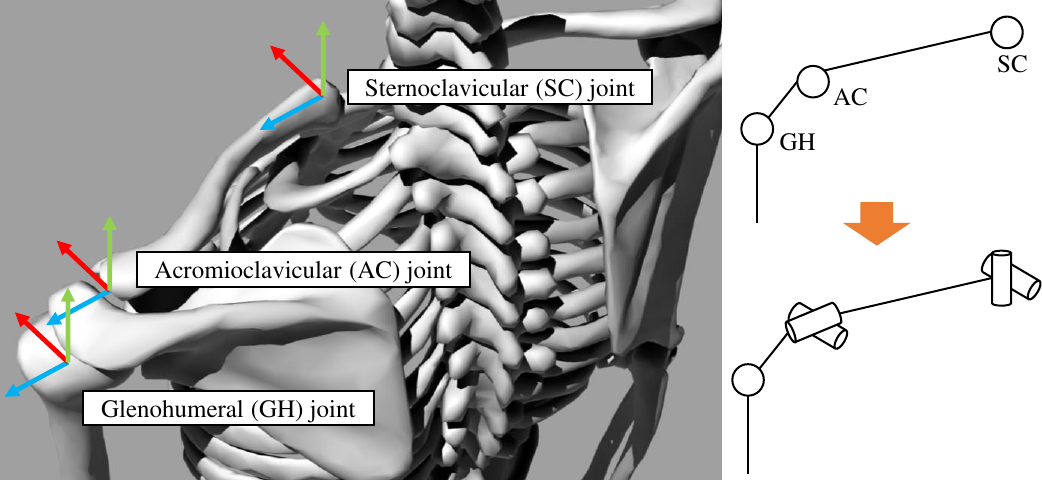}
  \caption{Details of the scapula structure. The right figure shows the change in DOFs arrangement.}
  \label{figure:kengoro-scapula-structure}
  \vspace{-1.0ex}
\end{figure}

\switchlanguage%
{%
  We will state the construction method of the polyarticular-JMM.
  We treat Kengoro as one example of a tendon-driven musculoskeletal humanoid, and show the relationship between the joints and muscles of Kengoro \cite{humanoids2016:asano:kengoro} in \figref{figure:kengoro-joint-muscle-table}.

  First, we choose joints to estimate joint angles for, and obtain the muscles that move the joints.
  Second, we pick up all the joints that these muscles can move.
  Finally, we construct JMM using these joints and muscles.
  For example, when we want to estimate the joint angles of the 3-DOF shoulder, at first, we pick up the muscles that can move the 3-DOF joint.
  The muscles include polyarticular muscles such as the biceps brachii, pectoralis major, latissimus dorsi, etc., so we pick up all the joints that these muscles can move.
  The joints include the elbow and scapula joints, and finally, we construct the JMM between 10 muscles and 10-DOF (the 3-DOF shoulder, 1-DOF elbow, and 6-DOF scapula).
  By using the same procedure, the polyarticular-JMM of the neck joint is the JMM between 10 muscles and 16-DOF.

  One of the most important reasons why the number of DOFs included in a polyarticular-JMM become large is that the number of DOFs of the scapula joint(the SC and AC joints), is 6, which is very large.
  It is clear that the construction of polyarticular-JMMs becomes easier by reducing the number of DOFs of the scapula.

  Therefore, we will consider the structure of the scapula first.
  The detailed structure of the scapula is shown in \figref{figure:kengoro-scapula-structure}.
  In the SC joint (roll, pitch, yaw) and AC joint (roll, pitch, yaw), the pitch of the SC joint and the yaw of the AC joint do not move very much.
  To be precise, we can prevent the scapula from coming off of the thorax by using the pitch of the SC joint or the pitch of the AC joint.
  Also, the yaw of the AC joint moves only about a maximum of 10 deg as shown in \cite{none:kapandji:upper}.
  From these facts, we believe that we can reduce the number of DOFs of the scapula joint.
  Even though this reduction is not completely accurate, because the motions of the scapula can be expressed roughly, we reduce the number of DOFs of the scapula joint to 4 to make the construction of polyarticular-JMMs easier in this study.
}%
{%
  最初に、関節-筋空間マッピングの構成方法について述べる。
  例として、腱悟郎\cite{humanoids2016:asano:kengoro}の関節と筋の関係を\figref{figure:kengoro-joint-muscle-table}に示し、これを例とすることで理解を進める。
  まず、関節角度推定を行いたい関節群を取り出し、その関節群を動作させることのできる筋群を得る。
  そして、その筋群が動かすことのできる関節群を取り出す。
  例を挙げると、まず、肩の3自由度の関節角度推定を行いたい場合には、その3自由度を動かすことのできる筋群を取り出す。
  その筋群には上腕二頭筋や大胸筋等の多関節が多数含まれており、その筋群が動かすことのできる関節群をさらに取り出す。
  これには、肘関節や肩甲骨が含まれており、最終的に10本の筋群と10自由度の関節群の対応である関節-筋空間マッピングが構成される。
  また、同様の理論で首の関節-筋空間マッピング作成には16自由度と10本の筋の対応が必要である。
  このように対応する関節の自由度数が多くなってしまう理由の最も大きな要因として、肩甲骨の自由度が6自由度と非常に大きいことが挙げられる。
  これを減らすことができれば、関節-筋空間マッピングが構成しやすくなることは明らかである。

  そこでまず、肩甲骨の構造について考察する。
  肩甲骨の詳細な構造は\figref{figure:kengoro-scapula-structure}のようになっている。
  この胸鎖関節のroll, pitch, yawと肩鎖関節のroll, pitch, yaw軸において、大きく動作しないものとして胸鎖関節のpitch, 肩鎖関節のyawが挙げられる。
  正確には胸鎖関節のpitchか肩鎖関節のpitchのどちらかで肩甲骨が胸郭から剥がれないようにすることは可能であり、また、肩鎖関節のyaw軸は\cite{none:kapandji:upper}にあるように最大で10度程度しか動作しない。
  よってこれらを肩甲骨の自由度から除くことを考える。
  これは完全に正しいわけではないが、このように考えても, 肩甲骨の動作を全体的に表現することができていることから、この自由度で十分であると今回は考えた。
  よって、肩甲骨の自由度を4自由度に減らすことができ、これによって関節-筋空間マッピングが構成しやすくなった。
  このように、非常に単純な話ではあるが、不必要な自由度群を削除してしまう、というのは重要である。
}%

\begin{figure}[htb]
  \centering
  \includegraphics[width=1.0\columnwidth]{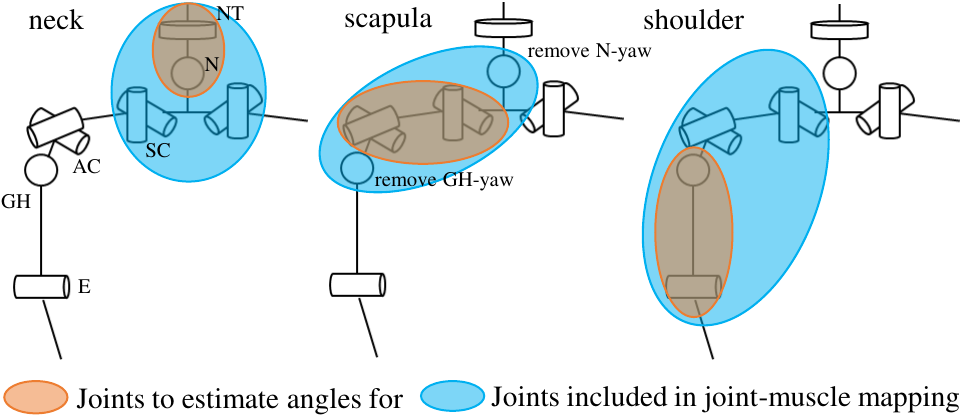}
  \caption{An example of joint-muscle mapping construction.}
  \label{figure:joints-grouping}
  \vspace{-1.0ex}
\end{figure}

\subsection{Dividing Joints and Muscles into Several Groups}
\switchlanguage%
{%
  By dividing joints and muscles into several groups and constructing a polyarticular-JMM for each group, the construction of whole body JMM can become easier.
  Although how the joints and muscles are grouped does not matter, in terms of maintenance, it is better to construct a few number of polyarticular-JMMs in which as many joints and muscles are included as to the limit of computational cost, than to construct a large number of small polyarticular-JMMs.
   does not matter
  As one example, we show the grouping of joints and muscles in upper limb of Kengoro in \figref{figure:joints-grouping}.
  Although we cannot divide joints and muscles simply due to the influence of polyarticular muscles such as the biceps brachii, pectoralis major, trapezius, and latissimus dorsi, we can divide JMM into small polyarticular-JMMs by permitting the duplication of DOFs across several polyarticular-JMMs.
  Also, in terms of computational cost, we limited the number of DOFs included in each polyarticular-JMM to 8.
  The result of the grouping is shown in \figref{figure:joints-grouping}.
  The polyarticular-JMM of the neck includes the roll and yaw DOFs of the right and left scapula AC joints.
  The trapezius and deltoid (front) cannot move the roll and pitch DOFs of the SC scapula joint very much, so we reduced these unnecessary DOFs as in the theory stated in \secref{subsec:reduce-scapula}.
  Also, the polyarticular-JMM of the scapula include the roll and pitch DOFs of the neck and shoulder.
  Like in the neck, we reduce the yaw DOF of the neck and shoulder from the scapula polyarticular-JMM, because muscles around the scapula do not move these DOFs very much.
  The polyarticular-JMM of the shoulder and elbow includes the roll and yaw of the AC scapula joint and the roll and pitch of the SC scapula joint.

  In general, we should set a limitation to the number of DOFs included in each polyarticular-JMM in terms of computational cost, and choose joints and muscles so as not to exceed this limit.
  In this JMM construction method, polyarticular-JMMs include DOFs duplicated among several polyarticular-JMMs, and we cope with this next.
}%
{%
  関節群と筋群を小分けし、いくつかの関節-筋空間マッピングに分けることで関節-筋空間マッピングの構成を楽にすることができる。
  これはどのように分けても構わないが、小さな関節-筋空間マッピングが多いよりは、保守管理の観点から、計算できるギリギリの大きさの関節-筋空間マッピングが少数あったほうが望ましい。

  腱悟郎の上半身を例に取って、グルーピングの例を示す(\figref{figure:joints-grouping})。
  上腕二頭筋や大胸筋、僧帽筋や広背筋等の多関節筋の影響によって、関節群と筋群をスッパリと分けることは出来ないが、含まれる関節群の重複を許すことで関節-筋空間マッピングを小分けにすることができる。
  また、計算量的な観点から一つの関節-筋空間マッピングに含まれる関節群を8までに抑えるような工夫をしている。
  その結果が\figref{figure:joints-grouping}であるが、首に関する関節-筋空間マッピングには左右の肩甲骨のAC関節のroll, yawが含まれている。
  またこの際、僧帽筋はSC関節のroll pitchを大きく動かすことはできないため、前節の考え方と同様に削除している。
  肩甲骨に関する関節-筋空間マッピングには首のroll, pitchとshoulderのroll pitchが含まれている。
  これも同様に、首のyawとshoulderのyawから肩甲骨周辺の筋は大きく影響を受けないため削除している。
  肩と肘に関する関節-筋空間マッピングには肩甲骨のAC-roll, yaw, SC-roll, pitchが含まれている。

  これはほんの一例であるが、計算量的な観点から関節-筋空間マッピングに含まれる関節群の個数限度を設定し、それを超えないように関節群・筋群を選ぶことが望ましい。
  しかしこの場合、関節-筋空間マッピング間で重複する関節群が含まれるため、これを次節で適切に処理する。
}%

\subsection{Formulation of a Joint Angle Estimation Method Using Polyarticular Joint-Muscle Mapping}
\switchlanguage%
{%
  In this subsection, we will discuss the formulation of the joint angle estimation method using polyarticular-JMM by modifying a previous work \cite{humanoids2015:okubo:muscle-learning}.
  In this method, there is a group of joints for which the angles need to be estimated, and a group of joints for which this is not necessary.
  For example in Kengoro, the polyarticular-JMM used to estimate the 4-DOF scapula joint angles includes two neck DOFs and two shoulder DOFs in addition to the 4-DOF scapula joint.
  The group of joints we want to estimate angles for, $\bm{\theta}_y$, is the 4-DOF scapula joint, and the group of joints we do not need to estimate angles for, $\bm{\theta}_n$, is the 4-DOF of the neck and shoulder.
  The formulation of the polyarticular-JMM is shown as below.
  \begin{align}
  \bm{l} = f(\bm{\theta} = \begin{pmatrix} \bm{\theta}_y \\ \bm{\theta}_n \end{pmatrix})
  \end{align}
  The state equation of EKF (\equref{equation:state-normal}) in the previous joint angle estimation method \cite{humanoids2015:okubo:muscle-learning} is expressed as below.
  \begin{align}
  \bm{\theta}_{k|k-1} = {\begin{pmatrix} \bm{\theta}_y \\ \bm{\theta}_n \end{pmatrix}}_{k|k-1} = \bm{\theta}_{k-1|k-1} + G^{+}(\bm{\theta}_{k-1|k-1})\delta\bm{z}_{k}
  \end{align}
  In this case, it is easy for $G^{+}(\bm{\theta}_{k-1|k-1})\delta\bm{z}_{k}$ to become unexpected, because $\bm{\theta}_{n, k|k-1}$ can be predicted using not only the displacement of the polyarticular muscles included in the polyarticular-JMM, but also other monoarticular muscles included in the other JMMs.
  Therefore, we modify the formulation as shown below, and remove the prediction of $\bm{\theta}_n$.
  \begin{align}
    \bm{\theta}_{k|k-1} = {\begin{pmatrix} \bm{\theta}_y \\ \bm{\theta}_n \end{pmatrix}}_{k|k-1} &= \bm{\theta}_{k-1|k-1}\\\nonumber&+ \begin{pmatrix} I_y \quad O_n \end{pmatrix}G^{+}(\bm{\theta}_{k-1|k-1})\delta\bm{z}_{k} \label{equation:poly-estimation-first}
  \end{align}
  Also, because $\bm{\theta}_n$ is the joint angles obtained by the joint angle estimation using the other polyarticular-JMMs, after the calculation of kalman filter,  $\bm{\theta}_{k|k}$ should be overwritten.
  If this is not done, $\bm{\theta}_n$ can become gradually larger or vibrate, and badly influence the estimation of $\bm{\theta}_y$, because $\bm{\theta}_{n, k|k} $ is calculated only from polyarticular muscles which cross other DOFs.
  So, we overwrite $\bm{\theta}_{n, k|k}$ as shown below.
  \begin{align}
    \bm{\theta}_{n, neck, k|k} &= \bm{\theta}_{y, neck, k|k} \\
    \bm{\theta}_{n, scapula, k|k} &= \bm{\theta}_{y, scapula, k|k} \\
    \bm{\theta}_{n, shoulder, k|k} &= \bm{\theta}_{y, shoulder, k|k} \label{equation:poly-estimation-last}
  \end{align}
  By these methods, we can estimate joint angles using polyarticular-JMM.
}%
{%
  本章では既存の手法\cite{humanoids2015:okubo:muscle-learning}を変更した、多関節筋の影響によって、含まれる関節群の重複する関節-筋空間マッピングらを用いた関節角度推定の定式化について議論をする。
  多関節筋を含む関節角度推定では、関節-筋空間マッピングの関節に、関節角度推定で求めたい関節群と求める必要がない関節群が存在する。
  例えば腱悟郎の場合、肩甲骨の関節角度推定は肩甲骨の4軸の関節角度を推定するが、肩甲骨の関節-筋空間マッピングには肩甲骨4軸以外に首の2軸と肩の2軸が含まれている。
  ここでは求めたい関節群$\bm{\theta}_y$は肩甲骨の4軸、求める必要がない関節群$\bm{\theta}_n$は首と肩の4軸となる。
  よって関節-筋空間マッピングの式は
  \begin{align}
  \bm{l} = f(\bm{\theta} = \begin{pmatrix} \bm{\theta}_y \\ \bm{\theta}_n \end{pmatrix})
  \end{align}
  となる。
  ここで、カルマンフィルタを用いた関節角度推定の状態方程式\equref{equation:state-normal}を$\bm{\theta}_y$と$\bm{\theta}_n$を用いて以下に示す。
  \begin{align}
  \bm{\theta}_{k|k-1} = {\begin{pmatrix} \bm{\theta}_y \\ \bm{\theta}_n \end{pmatrix}}_{k|k-1} = \bm{\theta}_{k-1|k-1} + G^{+}(\bm{\theta}_{k-1|k-1})\delta\bm{z}_{k}
  \end{align}
  この際に問題と成るのは$\bm{\theta}_{n, k|k-1}$は実際にはこの関節-筋空間マッピングに含まれる多関節筋のみから予測されるものではないため、$G^{+}(\bm{\theta}_{k-1|k-1})\delta\bm{z}_{k}$の値が飛びやすいということである。
  そこで、以下のように定式化を変更し、$\bm{\theta}_n$の予測をなくす。
  \begin{align}
  \bm{\theta}_{k|k-1} = {\begin{pmatrix} \bm{\theta}_y \\ \bm{\theta}_n \end{pmatrix}}_{k|k-1} = \bm{\theta}_{k-1|k-1} + \begin{pmatrix} I_y \quad O_n \end{pmatrix}G^{+}(\bm{\theta}_{k-1|k-1})\delta\bm{z}_{k} \label{equation:poly-estimation-first}
  \end{align}
  また、$\bm{\theta}_n$はこれとは別の関節-筋空間マッピングからの関節角推定で求まる部分であるため、カルマンフィルタの予測・更新が終わった後に求まった$\bm{\theta}_{k|k}$を上書く必要がある。
  これをしない場合、$\bm{\theta}_n$が少ない多関節の筋長変化から求まるため非常に不安定であり、値が徐々に大きくなったり振動したりしてしまうことが見受けられる。
  よって
  \begin{align}
    \bm{\theta}_{n, k|k, neck} &= \bm{\theta}_{y, k|k, neck} \\
    \bm{\theta}_{n, k|k, scapula} &= \bm{\theta}_{y, k|k, scapula} \\
    \bm{\theta}_{n, k|k, shoulder} &= \bm{\theta}_{y, k|k, shoulder} \label{equation:poly-estimation-last}
  \end{align}
  のように上書く。
  この方法によって、求めたい関節群以外も含んでしまう多関節筋を含む関節-筋空間マッピングの関節角度推定が可能となる。
}%

\section{Joint Angle Estimation using Only the Relative Changes in Muscle Lengths} \label{sec:4}
\switchlanguage%
{%
  In this section, we extend the joint angle estimation method using the absolute muscle lengths (absolute-JAE) \cite{humanoids2015:okubo:muscle-learning}, and propose a method that estimates the current joint angles using only the relative changes in muscle lengths (relative-JAE).
  In the EKF, absolute muscle lengths are needed for the observation equation (\equref{equation:observation-normal}).
  In this study, we modify the observation equation to a form that uses the nonlinear feature of JMM (we use only the muscle jacobian and the relative changes in muscle lengths), and does not use the absolute muscle lengths.
  Thus, the observation equation of EKF is changed as shown below.
  \begin{align}
    \delta\bm{\theta}_{k} &= \bm{\theta}_{k|k-1} - \bm{\theta}_{k-1|k-1} \\
    \delta\bm{z}_{k} &= G(\bm{\theta}_{k-1|k-1})\delta\bm{\theta}_{k} \label{equation:observation-relative}
  \end{align}
  The state equation is the same as \equref{equation:poly-estimation-first}.
  In this equation, we observe not the absolute muscle lengths $\bm{z}_{k}$ but the relative changes in muscle lengths $\delta\bm{z}_{k}$.
  We can obtain $\delta\bm{z}_{k}$ by multiplying the muscle jacobian $G$ and the difference between the previous and current joint angles.
  As described, by merely changing the observation equation, a joint angle estimation method that does not require the absolute muscle lengths is possible.
  Thus we do not need to calibrate muscle lengths in this formulation.
}%
{%
  本章では、大久保らの絶対的な筋長からの関節角度推定\cite{humanoids2015:okubo:muscle-learning}をシンプルに拡張し、相対的な筋長変化のみから絶対的な関節角度推定を行う手法を述べる。
  カルマンフィルタで絶対筋長が必要な部分は、更新における\equref{equation:observation-normal}の絶対筋長の観測の部分である。
  よって、これを絶対筋長を使わず、相対的な筋長のみを用いて、関節-筋空間マッピングの非線形性を利用した形に置き換えてあげることを考える。
  つまり、絶対筋長の誤差を使ってupdateするのではなく、相対的な筋長変化の誤差を使ってupdateすれば良いのである。
  故にカルマンフィルタの式は多少変更され、以下のようになる。
  \begin{align}
    \nonumber\mbox{Predict}\;\;\;\;\;\;\;\;\;\;\;\;& \\
    \bm{\theta}_{k|k-1} &= \bm{\theta}_{k-1|k-1} + G^{+}(\bm{\theta}_{k-1|k-1})\delta\bm{z}_k \label{equation:state-relative} \\
    P_{k|k-1} &= P_{k-1|k-1} + Q \\
    \nonumber\mbox{Update}\;\;\;\;\;\;\;\;\;\;\;\;& \\
    \delta\bm{\theta}_{k} &= \bm{\theta}_{k|k-1} - \bm{\theta}_{k-1|k-1} \\
    \bm{e}_{k} &= \delta\bm{z}_{k} - G(\bm{\theta}_{k-1|k-1})\delta\bm{\theta}_{k} \label{equation:observation-relative} \\
    H_{k} &= \left.\frac{\partial G}{\partial \bm{\theta}}\delta\bm{\theta}_{k}\right|_{\bm{\theta} = \bm{\theta}_{k|k-1}} \\
    S_{k} &= H_{k}P_{k|k-1}H^T_{k} + R \\
    K_{k} &= P_{k|k-1}H^T_{k}S^{-1}_{k} \\
    \bm{\theta}_{k|k} &= \bm{\theta}_{k|k-1} + K_{k}e_{k} \\
    P_{k|k} &= (I-K_{k}H_{k})P_{k|k-1} \label{equation:last-relative}
  \end{align}
  このように、観測方程式の形を変えるだけで、非線形性を利用した初期値の必要ない相対的な筋長のみを用いた関節角度推定が可能となる。

  本研究では、ニューラルネットワークを用いた方法においてこれを実装し、実験を行っている。
}%

\section{Experiments} \label{sec:5}
\switchlanguage%
{%
  As basic experiments, we validate that we can correctly construct the neck, scapula, and shoulder polyarticular-JMMs stated in \secref{sec:3}, and that we can estimate joint angles using polyarticular-JMMs in a simulation environment.
  Also, regarding the shoulder, we validate the effectiveness of the joint angle estimation method using the actual tendon-driven musculoskeletal humanoid, Kengoro.
  Next, we conduct experiments of joint angle estimation using only the relative changes in muscle lengths in a simulation environment.
  Also, we validate the effectiveness of the joint angle estimation method using the actual robot, Kengoro.
}%
{%
  まず、\secref{sec:3}で述べた関節-筋空間マッピングの構築方法によって首、肩甲骨、肩の関節-筋空間マッピングが計算量的に問題ない範囲で構築できること、また、シミュレーションによって多関節筋を含んだ多自由度の関節角度推定の検証を行う。
  また、肩に関して実機での関節角度推定を行いその有効性を確認する。
  次に、相対的な筋長のみを用いた関節角度推定について、シミュレーションにおける検証を行う。
  また、実機を用いた相対的な筋長のみを用いた関節角度推定の有効性の検証を行う。
}%

\subsection{Joint Angle Estimation Using Polyarticular Joint-Muscle Mapping}
\begin{figure}[t]
  \centering
  \includegraphics[width=0.9\columnwidth]{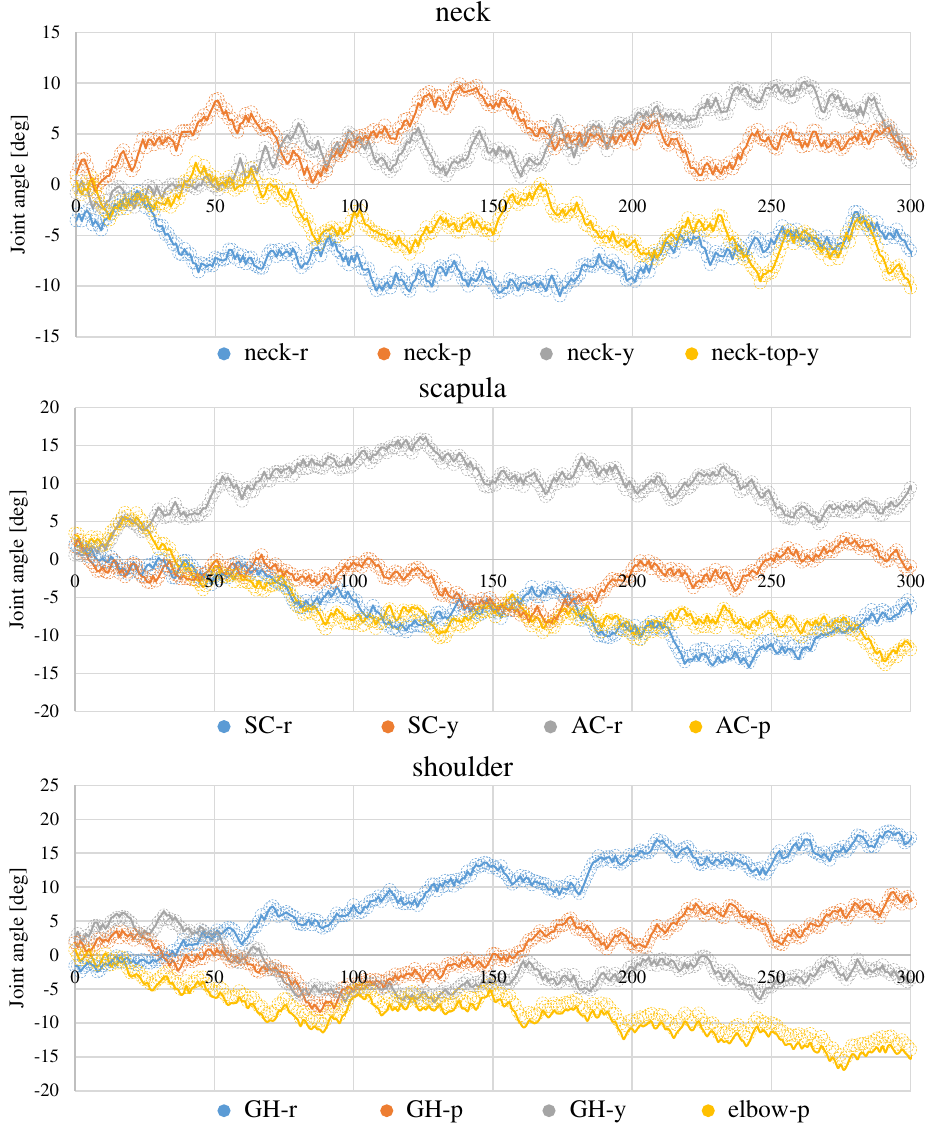}
  \caption{The result of joint angle estimation of the neck, scapula, and shoulder using the geometric model. The actual joint angles (circle) and the estimated joint angles (line) are almost the same.}
  \label{figure:joint-angle-estimation}
  \vspace{-1.0ex}
\end{figure}

\switchlanguage%
{%
  First, we will summarize the computational cost required to construct the neck, scapula, and shoulder polyarticular-JMMs stated in \secref{sec:3}.
  We show the conditions and actual calculation time required to construct each polyarticular-JMM in \tabref{table:comparison-joint-muscle}.
  The specifications of the machine used for this calculation are CPU:Intel(R) Core(TM) i7-5930K CPU @ 3.50GHz 12 Core, Memory:128GB.
  The data set was generated by setting $N$ from 5 to 9 (we used 5 regarding joints having small movable ranges, and 9 regarding joints having large movable ranges).
  The number of the data set is $N^D$, as stated in \secref{sec:3}.
  The number of DOFs $D$ and the number of muscles $M$ are the same as stated in \secref{sec:3}.
  The data set construction and polynomial regression were completed in a practical amount of time, and so the approach in this study proved to be effective.

  Second, we verified the joint angle estimation method using the polyarticular-JMMs.
  We used a geometric model in this experiment.
  We estimated joint angles from the displacement of muscle lengths when moving the joint angles of the model using random walk, and verified if the movement of the geometric model and estimated joint angles match.
  We show the result in \figref{figure:joint-angle-estimation}.
  The estimated joint angles match the actual movement, and we can see that the joint angle estimation method using polyarticular-JMM is feasible.
}%
{%
  まず、首・肩甲骨・肩のそれぞれについて関節-筋空間マッピングを構成する際の時間についてまとめる。
  \tabref{table:comparison-joint-muscle}にそれぞれのマッピング作成時の条件と所要時間を示す。
  使ったマシンのスペックはCPU:Intel(R) Core(TM) i7-5930K CPU @ 3.50GHz 12 Core, Memory:132GBである。
  データ数は各関節角度を5--9等分し(関節角度範囲が小さいものは5等分、大きなものは9等分)、している。
  これは\secref{sec:3}における$N^D$である。
  また、DOFとmuscle numは\secref{sec:4}でグルーピングしたものと同じものである。
  ここではデータ作成・多項式近似において現実的な時間で計算が完了しており、関節-筋空間マッピングのグルーピングの有効性が示されている。

  次に、多関節筋を含んだ多自由度の関節角度推定について検証を行う。
  この実験では幾何モデルのみを用いて実験を行い、幾何モデルを動作させたときの筋長の変化から関節角度を推定し、幾何モデルの動作と関節角度推定値がどの程度合っているかについて検証する。
  \figref{figure:joint-angle-estimation}にその結果を示すが、多関節筋を含んだ多数の自由度を含む関節-筋空間マッピングでも関節角度推定が可能なことがわかる。
}%

\begin{table}[t]
  \centering
  \caption{Computational cost to construct polyarticular-JMMs of upper limb of Kengoro (the neck, scapula, and shoulder).}
  \scalebox{1.0}{
    \begin{tabular}{l | r | r | r}
      \hline
      & neck & scapula & shoulder \\ \hline
      number of data ($N^D$) & 1071875 & 2205000 & 2646000 \\
      DOFs ($D$) & 8 & 8 & 8 \\
      number of muscles ($M$) & 10 & 8 & 10 \\
      data set construction time [sec] & 8579 & 16957 & 16840 \\
      polynomial regression time [sec] & 5514 & 12885 & 25287 \\\hline
    \end{tabular}
  }
  \label{table:comparison-joint-muscle}
\end{table}

\begin{figure}[t!]
  \centering
  \includegraphics[width=1.0\columnwidth]{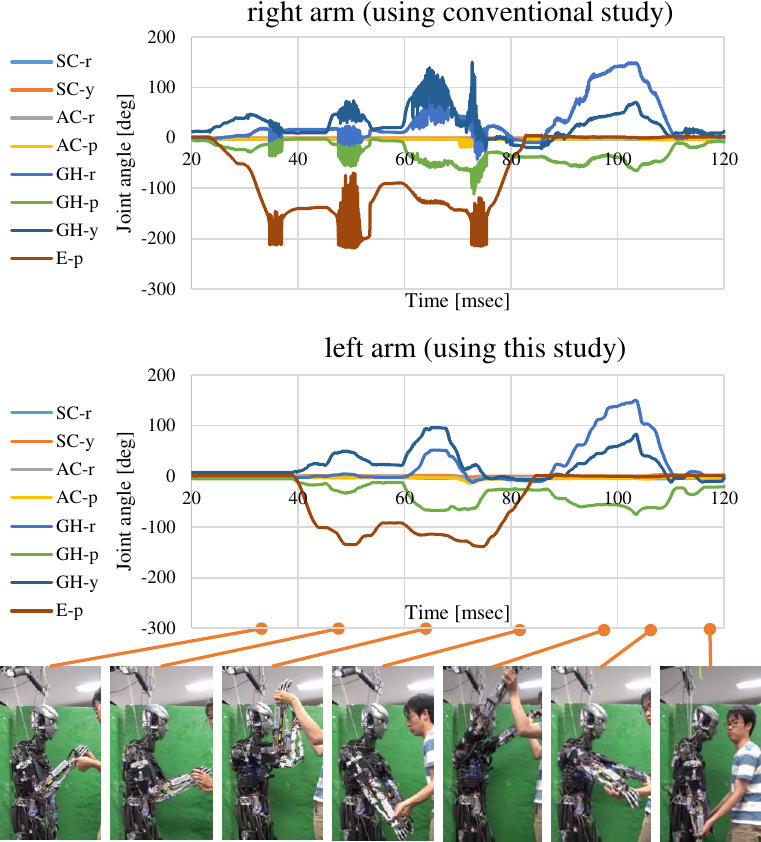}
  \caption{The result of joint angle estimation of the neck, scapula, and shoulder using the actual robot, Kengoro.}
  \label{figure:actual-joint-angle-estimation}
  \vspace{-1.0ex}
\end{figure}

\switchlanguage%
{%
  Third, we executed the joint angle estimation using the actual robot, Kengoro.
  We controlled Kengoro with constant tension control, moved Kengoro by applying force from the outside, and examined how the estimated joint angles changed.
  Because force is applied from the outside, it is easy for the estimation of joint angles to become incorrect by sudden noise, and we want to examine the robustness in this situation.
  We used the polyarticular-JMM obtained, stated above, and applied the conventional joint angle estimation method \cite{humanoids2015:okubo:muscle-learning} to the right arm and the joint angle estimation formulated in this study, which considers the influence of polyarticular muscles, to the left arm.
  The conventional joint angle estimation is the estimation in which \equref{equation:poly-estimation-first}--\equref{equation:poly-estimation-last} are not applied.
  First, we moved each elbow joint, and after that, moved both arms manually so that the joint angles will become almost the same.
  The result is shown in \figref{figure:actual-joint-angle-estimation}.
  While the estimated joint angles vibrated regarding the conventional joint angle estimation in the right arm, the estimated joint angles were stable and the changes were smooth regarding the joint angle estimation method of this study in the left arm.
  In the experiment, because the conventional joint angle estimation method does not share duplicated joint angles among several polyarticular-JMMs, the value of $\bm{\theta}_{n}$ became strange and influenced the value of $\bm{\theta}_{y}$.
  Also, because the vibration is large, it is difficult to remove using a simple low-pass filter.
}%
{%
  最後に実機実験としては、張力を一定に保つような制御を腱悟郎に施し、外部から人が腱悟郎の腕を動作させることで、関節角度推定がどのように推移するかを調べる。
  外部から人間が腱悟郎の体を動かすのは、急な外乱や大きな実機と幾何モデルの誤差などによって姿勢推定が崩れやすい難しい状況なため、よりロバストに姿勢推定ができているかを知るために、その状況を作り出す。
  また, dynamicに身体を動かすと, 本研究では考慮されていない筋張力の影響が大きく出てしまうため, quasi-staticにゆっくりと動作させる.
  関節-筋空間マッピングは上で得たものを用い、右腕には従来の関節角度推定を、左腕には本研究で定式化した多関節筋の影響を考慮した関節角度推定を入れた。
  従来の関節角度推定とは、\equref{equation:poly-estimation-first}--\equref{equation:poly-estimation-last}を適用しないような関節角度推定である。
  最初に右肘、左肘の順に関節を曲げ、その後右腕左腕が同じような関節角度になるように外から動作させた。
  結果は\figref{figure:actual-joint-angle-estimation}のようになっている。
  右腕に関しては\secref{sec:5}で述べたような理由から推定した角度が振動しているのに対して、左腕に関しては振動がなく、関節角度の遷移が滑らかである。
  これは、従来は推定された関節角度を関節-筋空間マッピング間で共有していないが故に、$\bm{\theta}_{n}$が意図しない値になり、その影響で$\bm{\theta}_{y}$の関節角度も影響を受けているからである。
  また, 振動が大きいため, 単純にローパス等を使った場合でも振動を取り除くことは難しい.
  それに対して, 本研究の手法では, 推定された関節角度を関節-筋空間マッピング間で共有しているため, 振動せず, 正しく関節角度推定が成功しており, 安定して動作している.
}%




\begin{figure*}[t]
  \centering
  \includegraphics[width=2.0\columnwidth]{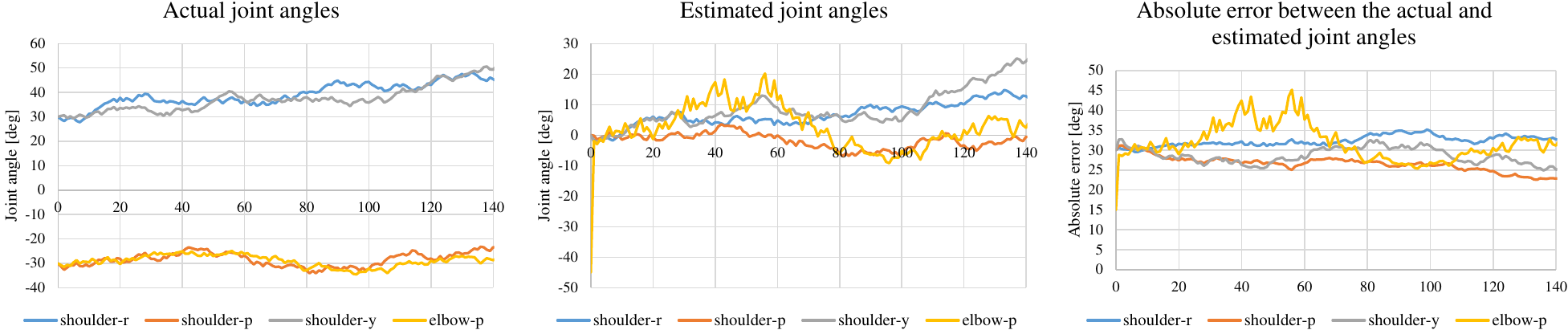}
  \caption{The result of joint angle estimation of the shoulder using absolute muscle lengths in the geometric model.}
  \label{figure:absolute-joint-angle-estimation}
\end{figure*}
\begin{figure*}[t]
  \centering
  \includegraphics[width=2.0\columnwidth]{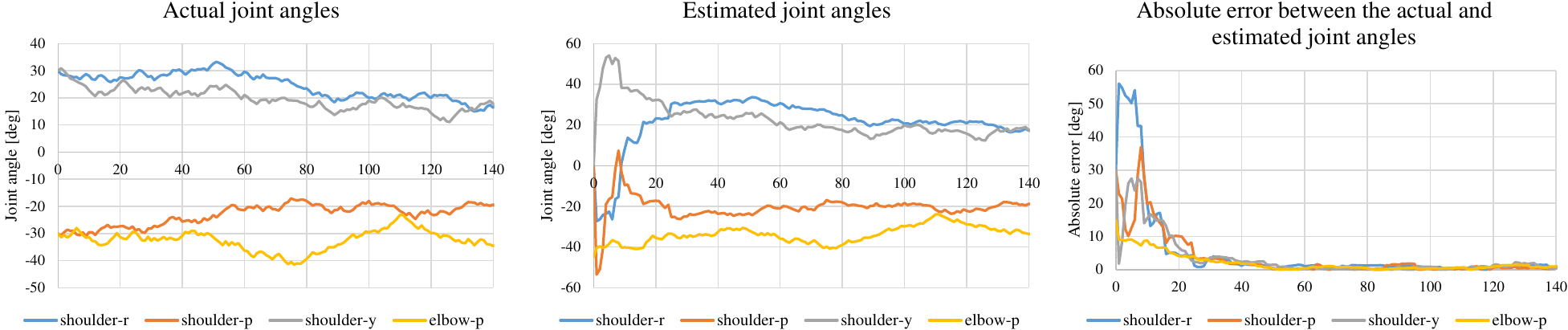}
  \caption{The result of joint angle estimation of the shoulder using only relative changes in muscle lengths in the geometric model.}
  \label{figure:relative-joint-angle-estimation}
  \vspace{-3.0ex}
\end{figure*}

\subsection{Joint Angle Estimation Using Only the Relative Changes in Muscle Lengths}
\begin{figure}[t]
  \centering
  \includegraphics[width=1.0\columnwidth]{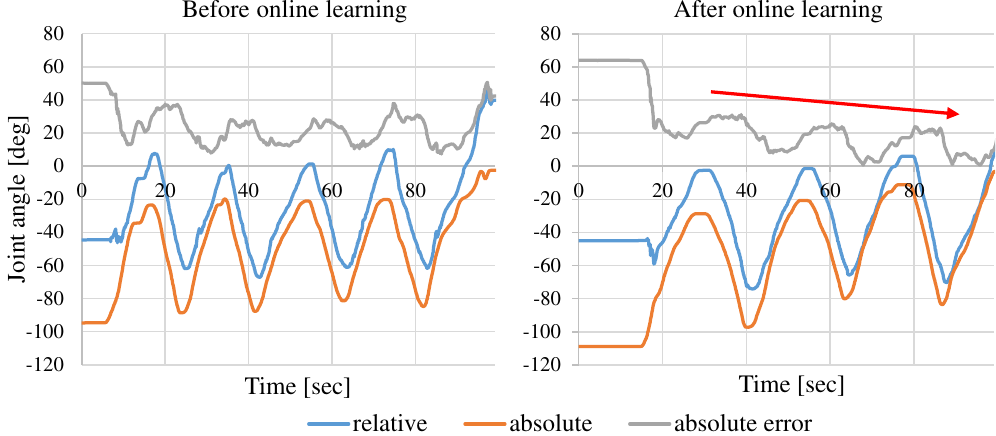}
  \caption{The result of the elbow joint angle estimation using only relative changes in muscle lengths in the actual robot Kengoro. Left graph is before online learning \cite{ral2018:kawaharazuka:vision-learning}, right graph is after online learning.}
  \label{figure:actual-elbow}
  \vspace{-1.0ex}
\end{figure}

\begin{figure}[t]
  \centering
  \includegraphics[width=1.0\columnwidth]{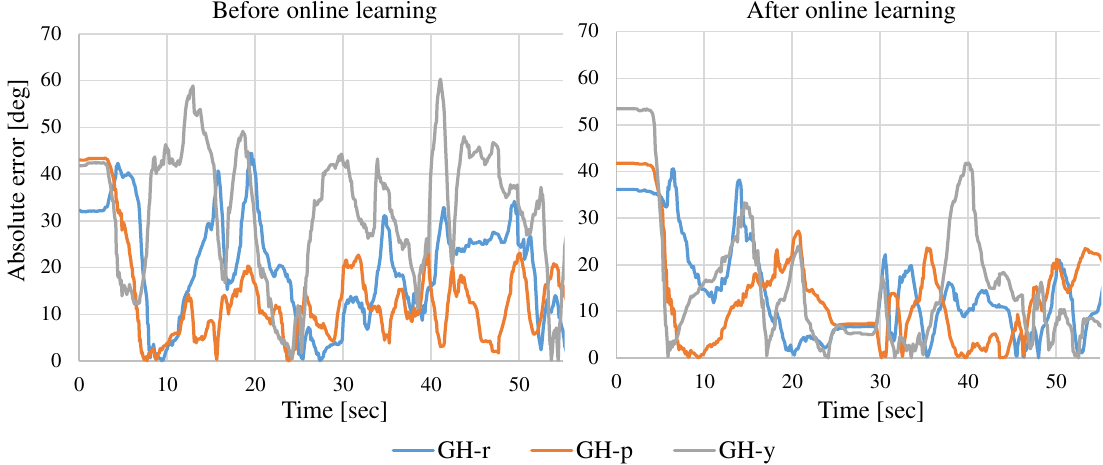}
  \caption{The result of the shoulder joint angle estimation using only relative changes in muscle lengths in the actual robot Kengoro. Left graph is before online learning \cite{ral2018:kawaharazuka:vision-learning}, right graph is after online learning.}
  \label{figure:actual-shoulder}
  \vspace{-1.0ex}
\end{figure}

\switchlanguage%
{%
  In this section, we used the formulation in \secref{sec:3} and \secref{sec:4}, but unlike in the previous subsection, we used JMM expressed not by polynomials but by a NN.
  This is because JMM obtained using a geometric model is different from the JMM of the actual robot, and we need to make them more similar by the online learning of JMM \cite{ral2018:kawaharazuka:vision-learning} using the sensor information of the actual robot.

  First, we conducted an experiment of relative-JAE stated in \secref{sec:4} in a simulation environment using the geometric model of Kengoro.
  Regarding the shoulder and elbow (GH-r, GH-p, GH-y, E-p), we conducted this experiment in an incorrectly calibrated situation in which the zero values of muscle lengths are calibrated when the joint angles are (30, -30, 30, -30) [deg].
  We began the experiment when the initial estimated joint angles were (0, 0, 0, -45) [deg] (because the maximum joint angle of the elbow is 0 deg, we offsetted it to -45 deg), moved each joint angle of the geometric model by random walk, and verified the transition of the estimated joint angles.
  We show the result of the conventional joint angle estimation (\equref{equation:state-normal}--\equref{equation:observation-normal}) in \figref{figure:absolute-joint-angle-estimation}.
  Because muscle lengths in JMM are 0 when joint angles are (0, 0, 0, 0) [deg], if joint angles at the muscle length calibration is misaligned at (30, -30, 30, -30) [deg], the estimated joint angles are unstable like in the center graph of \figref{figure:absolute-joint-angle-estimation}.
  Also, as shown in the right graph of \figref{figure:absolute-joint-angle-estimation}, the difference between the actual and estimated joint angles did not become smaller and the difference of about 30 deg did not change to a great extent.
  Next, we show the result of relative-JAE (\equref{equation:observation-relative}) in \figref{figure:relative-joint-angle-estimation}.
  In this method, because joint angles are estimated by using only the relative changes in muscle lengths and the nonlinear feature of JMM, while the conventional joint angle estimation could not decrease the difference between the actual and estimated joint angles when the joint angles at the calibration were incorrect, this method can estimate joint angles correctly.
  As shown in the right graph of \figref{figure:relative-joint-angle-estimation}, after random walk, the estimated joint angles became almost the same as the actual joint angles.
  Thus, we validated the effectiveness of this relative-JAE.

  Second, we conducted experiments of relative-JAE using the actual robot Kengoro.
  We estimated only the elbow joint angle at first.
  The initial value of the estimated joint angle is -45 deg, and we incorrectly calibrated muscle lengths to 0 when the joint angle is about -100 deg.
  To evaluate the correctness of the estimated joint angles, we preserved the muscle lengths when we calibrated correctly at 0 deg, and we estimated the elbow joint angle using the conventional absolute-JAE at the same time as relative-JAE.
  If the two estimated joint angles are the same (we call the difference between the two estimated joint angles JAEs-error), we can verify the effectiveness of relative-JAE in the actual robot.
  The accuracy of the absolute-JAE is evaluated in \cite{humanoids2015:okubo:muscle-learning}, so we only need to evaluate the JAEs-error, and not the difference between the estimated joint angles by relative-JAE and the actual joint angles.
  The result is shown in the left of \figref{figure:actual-elbow}.
  The JAEs-error became small to some extent (10--40 deg) during some exercises, but the result was not as good as in a simulation environment, because there is a model error between the actual robot and its geometric model.
  We show the result when we execute the antagonism updater of \cite{ral2018:kawaharazuka:vision-learning} for about 1 minute in the right of \figref{figure:actual-elbow} to correct antagonism of muscles.
  The JAEs-error became smaller (0--20 deg) than before learning.

  Finally, we conducted experiments of relative-JAE using the shoulder of Kengoro.
  We calibrated muscle lengths to 0 in the situation that the shoulder joint angles (GH-r, GH-p, GH-y) were (40, -40, 40), and began the experiment by setting the initial estimated joint angles to (0, 0, 0) [deg].
  We show the result in the left of \figref{figure:actual-shoulder}.
  The JAEs-error did not become smaller than in the elbow experiment, and the error fluctuated.
  Like the elbow experiment, we executed the antagonism updater of \cite{ral2018:kawaharazuka:vision-learning} and conducted the shoulder experiment again.
  The result is shown in the right of \figref{figure:actual-shoulder}.
  Although the JAEs-error became smaller, the fluctuation of error remained to some extent.

  As a result, learning of JMM using the actual robot can make relative-JAE stable and precise.
  However, the model error between the actual robot and its geometric model could not be modified completely, so the results were not as satisfactory as those from the experiments in a simulation environment.
  Therefore, we need to consider a method to make the JMM similar to the one in the actual robot and the formulation of a better joint angle estimation method.
}%
{%
  本節では、3節, 4節のformulationを用いているが、関節-筋空間マッピングは前節とは異なり、多項式近似ではなくニューラルネットワークを使ったものを使用している。
  これは、幾何モデルから得られた関節-筋空間マッピングは実機と大きく異なり, 実機オンライン学習\cite{ral2018:kawaharazuka:vision-learning}等によって、関節-筋空間マッピングを実機により近づけることが必要だからである.
  ネットワークのサイズ等は\cite{ral2018:kawaharazuka:vision}と同じものを使用している.

  まず、\secref{sec:4}で述べた相対的な筋長のみを用いた関節角度推定をシミュレーションにおいて試す。
  肩と肘(GH-r, GH-p, GH-y, E-p)に関して、ロボットを(30, -30, 30, -30)[deg]の関節角度で筋長の初期値を0にキャリブレーションした状態から始める。
  関節角度推定の値が最初は(0, 0, 0, -45)[deg]であるときに(肘は0[deg]が関節角度限界のため、初期値を-45[deg]までoffsetしている)から始め、ロボットにおけるそれぞれの関節をランダムウォークさせたときに関節角度推定がどのように遷移するかについて考察を行う。
  まず、通常の絶対筋長を用いた関節角度推定(\equref{equation:state-normal}--\equref{equation:last-normal})を行った場合の結果を\figref{figure:absolute-joint-angle-estimation}に示す。
  関節-筋空間マッピングは関節角度が(0, 0, 0, 0)[deg]の時が長さ0のため、(30, -30, 30, -30)[deg]のように、ズレた場所でキャリブレーションがなされていると推定値は\figref{figure:absolute-joint-angle-estimation}の中図のように非常に不安定である。
  また、右図のように実際のロボットの関節角度と関節角度推定値の差は絶対筋長が間違っているために縮まらず、最初の30[deg]程度のズレから変化していない。
  次に、本研究で開発した、相対筋長のみを用いた関節角度推定(\equref{equation:state-relative}--\equref{equation:last-relative})を行った場合の結果を\figref{figure:relative-joint-angle-estimation}に示す。
  ここでは、用いるのは相対的のみから非線形性を用いて関節角度推定を行っているため、絶対筋長を用いた関節角度推定では打ち消すことの出来ない実際のロボットの関節角度と関節角度推定値の差分を打ち消すことができる。
  \figref{figure:relative-joint-angle-estimation}の右図では、少しの試行で関節角度推定値が実際の関節角度とほとんど同じになっていることがわかる。
  このように、本研究で提案した手法が正しく動作することがわかる。

  次に、この相対的な筋長のみを用いた関節角度推定を実機において試す。
  まずは肘のみを動かして関節角度推定を行った場合を示す。
  関節角度推定の初期値は同様に-45[deg]とし、-100[deg]程度肘を曲げた状態で筋長の初期値を0にキャリブレーションした状態から始める。
  評価のため、絶対筋長(関節角度を全て0としたときを基準とした筋長)も取っておき、同時に、その絶対筋長を用いた関節角度推定を走らせる。
  この同時に走らせた相対的な筋長のみを用いた関節角度推定と絶対筋長を用いた関節角度推定の値が同じになれば、実機に置いて相対的な筋長のみで関節角度推定が行えることになる。
  その結果は\figref{figure:actual-elbow-basic}のようになった。
  関節-筋空間マッピングはロボットの幾何モデルから得られており、実機とは誤差があるため、シミュレーションのようには上手く行っていないのがわかる。
  relativeとabsoluteの関節角度推定値の差は10--30[deg]程度であり、その誤差はなかなか縮まらない。
  そこで、ニューラルネットワークによって構築した関節-筋空間マッピングの良さである実機における関節-筋空間マッピングのオンライン学習\cite{ral2018:kawaharazuka:vision-learning}を1分程度行った場合の結果を\figref{figure:actual-elbow-learned}に示す。
  このとき、relativeとabsoluteの誤差は学習しないときよりも小さくなっているのがわかる。

  最後に、同様の実験を肩関節を動かして行う。
  関節角度推定の初期値は(GH-r, GH-p, GH-y) = (0, 0, 0)[deg]とし、(40, 40, 40) [deg]程度に動かした状態で筋長の初期値を0にキャリブレーションした状態から始める。
  その結果を\figref{figure:actual-shoulder-basic}に示す。
  肘よりも複雑な肩関節では、relativeとabsoluteの間の誤差は縮まらず、振動がちになってしまっているのがわかる。
  同様に、関節-筋空間マッピングのオンライン学習を1分間程度行った際の結果を\figref{figure:actual-shoulder-learned}に示す。
  学習を行わない場合に比べて誤差も小さく安定してはいるが、それでも誤差は残り、振動も残っているのがわかる。

  このように、実機による関節-筋空間マッピングの学習によって関節-筋空間マッピングを実機に近づけることによって関節角度推定を安定に、そして正しくすることができる。
  しかし、現状ではまだ実機と幾何モデルの間の誤差を完全に修正することができていないため、実機では、シミュレーションで得られるような良い結果は得られていない。
  よって今後、より関節-筋空間マッピングを実機に近づける方法、または、より良い関節角度推定の定式化を考える必要があると思われる。
}%

\section{CONCLUSION} \label{sec:6}
\switchlanguage%
{%
  In this study, we discussed a method of joint angle estimation from muscle lengths in tendon-driven musculoskeletal humanoids, which include complex structures such as polyarticular muscles and the scapula, and an extended joint angle estimation method using only the relative changes in muscle lengths.
  Regarding the former, in previous joint angle estimation methods using data structures such as the data table, polynomials, and neural network, we stated that joint-muscle mapping (JMM), which expresses the nonlinear relationship between joint angles and muscle lengths, is important.
  Moreover, to construct the JMM, in terms of computational cost, we need to divide joints and muscles into several groups (polyarticular-JMMs) and formulate a joint angle estimation method that shares the estimated joint angles among groups while considering the effect of polyarticular muscles.
  Regarding the latter, we proposed a joint angle estimation method using only the relative changes in muscle lengths as an extension of the method using absolute muscle lengths.
  We considered the joint angle estimation of the current joint angles using the nonlinear feature of JMM by using not merely JMM but the differentiated value of JMM in the observation equation of EKF.
  Finally, we conducted experiments using these methods in simulation and actual environments, and verified the effectiveness.
  Regarding the latter, although the method functioned in a simulation environment or for the simple elbow joint of the actual robot, the estimated joint angles of the actual robot often diverged due to the model error between the actual robot and its geometric model, and so we need to consider a better learning framework to make the geometric model closer to the actual robot.

  In future works, we would like to improve stability and precision of the joint angle estimation method using only the relative changes in muscle lengths.
  Also, although we divided joints and muscles into several groups manually in this study, we would like to do it automatically.
}%
{%
  本研究では、多関節筋や肩甲骨等の複雑な構造を含む場合の筋長からの関節角度推定手法、そして、その発展となる相対的な筋長の変化のみを使った関節角度推定手法、という二つについて議論した。
  まず前者については、現在までに開発されているテーブルサーチや多項式近似、ニューラルネットワークを用いた関節角度推定全てについて関節空間から筋空間への写像を表現する関節-筋空間マッピングが重要であることを述べた。
  その上で、それを構成するためには計算量的観点から、関節群・筋肉群をグループに分ける必要があり、多関節筋の影響を考慮してそれらのグループ間で推定された関節角度を共有するという定式化を行った。
  また後者については、絶対的な筋長を用いた方法のシンプルな拡張として、相対的な筋長の変化のみから関節角度推定を行う手法を考案した。
  拡張カルマンフィルタの更新式として関節-筋空間マッピングではなく、その微分を用いることで、関節-筋空間の非線形性を用いて絶対的な関節角度を推定する方法を編み出した。
  最後に、これらに関してシミュレーションと実機において実験を行い、その有効性を示した。
  後者に関しては、シミュレーション環境や実機においても単純な関節であれば成功するものの、実機となると実機-モデル間の大きな誤差によって値が発散する場合が見受けれれたため、モデルを実機にさらに近づけるための学習的枠組みの必要性を感じた。

  今後は、相対的な筋長変化のみを用いた関節角度推定の安定化・高精度化を行っていきたい。
  また、今回関節群・筋群のグループ化を行ったが、これらを自動で行う等の試みを行いたい。
}%

\section*{Acknowledgement}
The authors would like to thank Yuka Moriya (Ochanomizu University) for proofreading this manuscript.

{
  \bibliographystyle{IEEEtran}
  \bibliography{main}
}

\end{document}